\documentclass{article}

\usepackage[preprint]{neurips_data_2024}

\usepackage[utf8]{inputenc} 
\usepackage[T1]{fontenc}    
\usepackage{hyperref}       
\usepackage{url}            
\usepackage{booktabs}       
\usepackage{amsfonts}       
\usepackage{nicefrac}       
\usepackage{microtype}      
\usepackage{xcolor}         
\usepackage{graphicx}       
\usepackage{pdflscape}
\usepackage{float}
\usepackage{booktabs}
\usepackage{array}
\usepackage{multirow}
\usepackage{geometry}
\usepackage{makecell}
\usepackage{amssymb}
\usepackage{relsize}
\usepackage{subcaption}
\usepackage{amsmath}
\usepackage{ulem}
\usepackage{inconsolata}

\usepackage{adjustbox}
\usepackage{array}

\newcommand*{\escape}[1]{\texttt{\textbackslash#1}}
\title{\textsc{GameBench}: Evaluating Strategic Reasoning Abilities of LLM Agents}

\author{%
  Anthony Costarelli\thanks{Equal contribution. Correspondence to: acostarelli@olin.edu} \\
  Olin College of Engineering \\
  \And
  Mat Allen\footnotemark[1] \\
  Independent \\
  \And
  Roman Hauksson\footnotemark[1] \\
  University of Texas at Dallas \\
  \And
  Grace Sodunke\footnotemark[1] \\
  University of Oxford \\
  \And
  Suhas Hariharan \\
  University College London \\
  \And
  Carlson Cheng \\
  Independent
  \And
  Wenjie Li \\
  ShanghaiTech University
  \And
  Joshua Clymer \\
  Columbia University
  \And
  Arjun Yadav \\
  University of Manchester
}

\begin{document}

\maketitle

\begin{abstract}
Large language models have demonstrated remarkable few-shot performance on many natural language understanding tasks. Despite several demonstrations of using large language models in complex, strategic scenarios, there lacks a comprehensive framework for evaluating agents’ performance across various types of reasoning found in games. To address this gap, we introduce \textsc{GameBench}, a cross-domain benchmark for evaluating strategic reasoning abilities of LLM agents. We focus on 9 different game environments, where each covers at least one axis of key reasoning skill identified in strategy games, and select games for which strategy explanations are unlikely to form a significant portion of models' pretraining corpuses. Our evaluations use GPT-3 and GPT-4 in their base form along with two scaffolding frameworks designed to enhance strategic reasoning ability: Chain-of-Thought (CoT) prompting and Reasoning Via Planning (RAP). Our results show that none of the tested models match human performance, and at worst GPT-4 performs worse than random action. CoT and RAP both improve scores but not to comparable human levels. Benchmark code is available at \url{https://github.com/Joshuaclymer/GameBench}.
\end{abstract}
\begin{figure}[!h]
\begin{subfigure}{0.5\textwidth}
\includegraphics[width=1\linewidth]{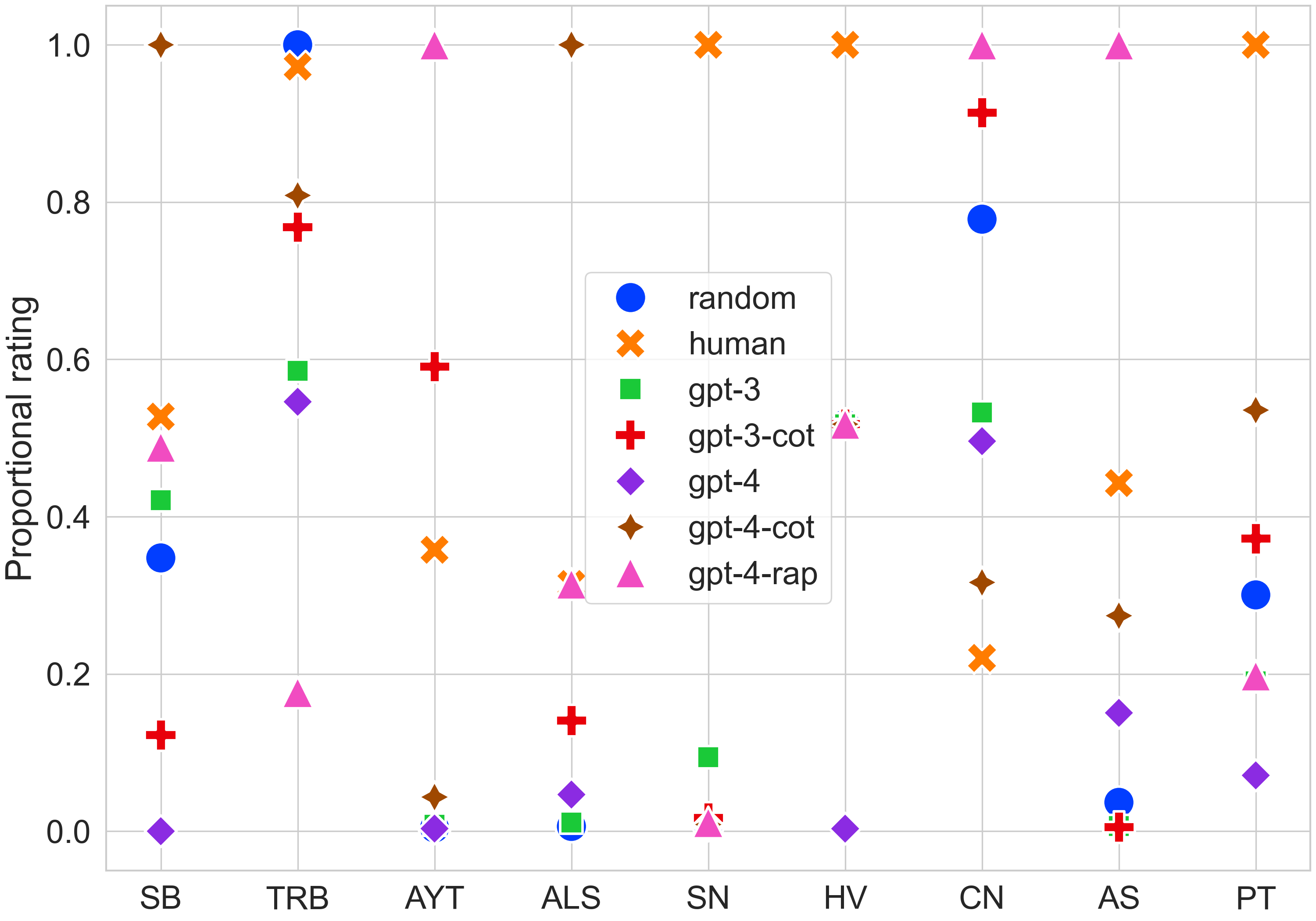}
\caption{Agent ratings per-game as proportion of best rating}
\label{fig:rating_scatter}
\end{subfigure}
\begin{subfigure}{0.5\textwidth}
\includegraphics[width=1\linewidth]{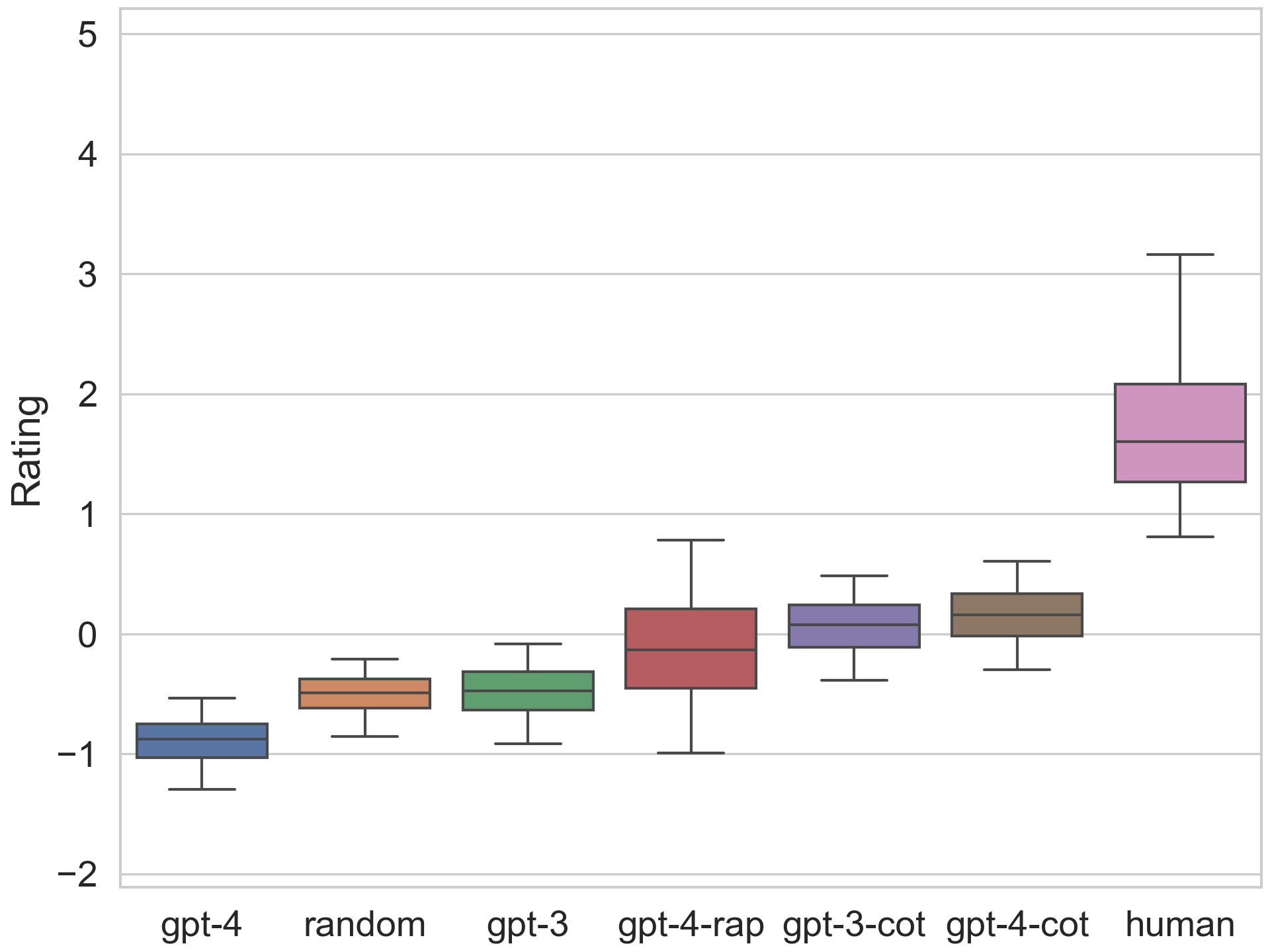}
\caption{Agent ratings overall (bootstrapped)}
\label{fig:subim2}
\end{subfigure}
\caption{\textbf{Rating data} With CoT scaffolding, GPT-4 is the best reasoner below only the human baseline, achieving the best LLM performance on \texttt{Sea Battle} and \texttt{Pit}. But without, it performs worse than even the random baseline due to its exceedingly low rating on \texttt{Sea Battle}. The state-of-the-art RAP scaffolding doesn't provide as much of an improvement to GPT-4 as CoT does. Looking at the top line of Figure \ref{fig:rating_scatter} reveal the best agent in each game. come from exponential Bradley–Terry model. See section \ref{rating_model} for details. The whiskers represent 90\% CIs computed from our bootstrapping process formalized in \ref{rating_model}. ALS = Air, Land, Sea; ARC = Arctic Scavengers; AYT = Are You the Traitor?; CN = Codenames; HV = Hive; PT = Pit; SN = Santorini; TRB = Two Rooms and a Boom; SB = Sea Battle.}
\label{fig:image2}
\end{figure}

\section{Introduction}
Capabilities of large language models have seen rapid progress, enabling LLMs to be used in agentic tasks \citep{schick2023toolformer} \citep{agent_gpt} \citep{auto_gpt}. This presents opportunities for LLM-based tools to assist humans in several domains, such as API usage \citep{li2023apibank}, web browsing \citep{schick2023toolformer} and coding \citep{kazemitabaar2023novices}. Recent benchmarks have been introduced for evaluating performance on real-world agent tasks \citep{wang2024mmlupro}, \citep{liu2023agentbench}, \citep{METR}, \citep{mialon2023gaia}, with some focused on reasoning \citep{sawada2023arb} and games \citep{lin2023agentsims}. However, these existing benchmarks are oriented to practical, in-distribution knowledge, which can quickly become saturated with better models.

In particular, strategic reasoning is an agentic task that is important for generalising to new contexts, as it involves optimising for an objective in the face of possibly divergent interests of others, where incentives may not be fully known \citep{gandhi2023strategic}. Prior work on reasoning scaffolds also shows that language models have potential to grasp reasoning skills across scenarios \citep{wei2022chain, hao2023reasoning}. Hence, a strategic reasoning benchmark for LLMs, that is inherently multi-agent, would be difficult to saturate. Furthermore, games exemplify environments for demonstrating strategic behaviour in both humans and AI agents, as seen in the well known examples of Chess \citep{silver2017mastering} and Go \citep{2016silveralphago}. Hence evaluating LLMs on several types of reasoning behaviours would present a comprehensive, fine-grained benchmark. As such, we introduce \textsc{GameBench}: a multi-player, cross-domain framework for evaluating strategic reasoning in LLM agents using games. We focus on both discrete and open-ended action spaces, across the reasoning domains of abstract strategy; non-deterministic outcomes; hidden information; language communication; social deduction and cooperation between players. By selecting for games without published strategy guides to our knowledge, we ensure that game-specific strategy has been sufficiently out-of-distribution in pretraining data. See Table \ref{game-properties-table} for a complete list of games and game properties.

The benchmark consists of obscure board games, card games, and social deception games. We evaluate \texttt{gpt-3.5-turbo-1106} (GPT-3) and \texttt{gpt-4-1106-preview} (GPT-4) along with the CoT \citep{wei2022chain} and RAP \citep{hao2023reasoning} scaffolding techniques, by playing them against each other, a random-action-selector baseline, and a human baseline. We conducted a literature review and identified RAP to be the state-of-the-art scaffolding that fit the parameters of our benchmark, i.e. each agent has access to the same game state information and no agent can peek at future states. Agents are rated using the exponential Bradley–Terry model \citep{bradleyterry}. This has useful advantages over the typical Elo system \citep{elo1967proposed}, such as its assumption that each agent’s ability is fixed and will not change between matches.

Our results show that CoT-augmented and RAP-augmented models demonstrate superior strategic superior to the random baseline; that GPT-3 matches the random baseline; that GPT-4  performs worse than the random baseline; and that the human baseline performs superior to all.

With this benchmark, we propose a means to measure the strategic reasoning abilities of LLM agents in diverse game environments. Our contributions are as follows:
\begin{itemize}
    \item \textbf{\textsc{GameBench}}, the first benchmark to capture both cross-domain and out-of-distribution strategic reasoning for comparison across multiple agents.
    \item \textbf{Empirical results} on GPT-3 and GPT-4, demonstrating the effects of Chain-of-Thought scaffolding and the state-of-the-art scaffolding.
\end{itemize}

\section{Related works}

\textbf{LLM agents playing games} Using games to evaluate LLMs has significant precedent in previous research. Some studies evaluate models using single strategic tasks or games, such as Minecraft \citep{Wang2023DescribeEP, Zhu2023GhostIT}, Diplomacy \citep{Bakhtin2022HumanlevelPI}, Avalon \citep{Light2023AvalonBenchEL}, and Werewolf \citep{Xu2023ExploringLL}. Other benchmarks \citep{Wu2023SmartPlayA, Liu2023AgentBenchEL} capture a more comprehensive picture by using suites of multiple tasks or games to evaluate LLMs as intelligent agents. However, the tasks represented in these benchmarks don't involve interaction with other agents, so they don't reflect strategic reasoning as defined in this work.

\textbf{Game-theoretic scenarios} Several benchmark suites focus on common game theory scenarios, such as auctions \citep{Chen2023PutYM, Mao2023ALYMPICSLA}, matrix games like Prisoner's Dilemma \citep{Akata2023PlayingRG, Gandhi2023StrategicRW}, and negotiation \citep{Abdelnabi2023LLMDeliberationEL, Gandhi2023StrategicRW}. While they do involve multi-agent interaction and are useful for testing models' strategic reasoning ability, our benchmark focuses on more complex games that aren't as frequently studied as these game theory scenarios. Given no major strategy guides or forums dedicated to these games, we believe there is less documentation on optimal strategies for them present in LLMs' training corpuses.

\textbf{Dialogue-based games} Some benchmarks employ dialogue-based games that are less well-documented on the internet: \citet{agashe2024llmcoordination} and \citet{Chalamalasetti2023clembenchUG} use novel cooperative dialogue games, and \citet{Qiao2023GameEvalEL} uses two social deduction games and one word guessing game. However, our benchmark aims to evaluate LLMs' strategic reasoning ability not only in cooperative and conversational environments, but competitive, spatial, and non-deterministic ones as well.

\textbf{Diverse multi-agent game suites} The benchmarks most similar to ours employ diverse suites of complex multi-agent games, including conversational, board, and card games \citep{Chen2024LLMArenaAC, Duan2024GTBenchUT, Abdulhai2023LMRLGB, Xu2023MAgICIO}. However, many of the included games are either commonly found on the internet, such as TicTacToe, Poker, and Connect Four, or common game-theoretic scenarios, as discussed previously. These games are not as out-of-distribution as desired.

In summary, we build upon previous work by introducing a diverse suite of multi-agent games to evaluate the strategic reasoning ability of LLMs as agents. Our benchmark is characterized by its inclusion of complex games that span a range of game characteristics and are not likely to be well-represented in LLMs' pretraining corpuses.

\section{\textsc{GameBench}}
\label{gamebench}

In Section \ref{agent_development} we discuss our reasoning behind our selection of agents and scaffolds. In Section \ref{game_selection} we describe our methodology for selecting suitable games. In Section \ref{api} we describe the agent and game interfaces. In Section \ref{rating_model} we introduce our rating model and formalize our process for calculating ratings.

\label{methodology}

\subsection{Agent and scaffolding selection}
\label{agent_development}

We benchmark GPT-3 (\texttt{gpt-3.5-turbo-1106}) and GPT-4 (\texttt{gpt-4-1106-preview}) due to their size, mainstream popularity, and convenient public API. We include these base models as well as several black-box scaffolding interventions in order to measure the relative effects these scaffolding interventions have on improving the reasoning abilities of the base models. We selected Chain-of-Thought \citep{wei2022chain} prompting for its ubiquity and Reasoning-via-Planning \citep{hao2023reasoning} for its state-of-the-art status. We also include a random-action-selecting agent as baseline of no strategic reasoning ability, and a human agent as a baseline of progress towards human-level strategic reasoning.

For more details about agent implementation, see Appendix \ref{implementation}.

\subsection{Game selection}
\label{game_selection}

In order to evaluate a broad range of cognitive skills associated with strategic reasoning, we curated a diverse set of games featuring abstract strategy, non-deterministic outcomes, hidden information, language communication, social deduction and bluffing, and cooperation between players. A breakdown of which games had these features can be found in Table \ref{game-properties-table}.

Using these categories, we then filtered for games unlikely to be significantly represented in LLMs' pretraining data, to evaluate the models' out-of-distribution reasoning abilities. Two key criteria were (a) excluding games with dedicated online forums discussing improvement strategies, as well as (b) excluding games with published strategy guides. After finalizing the selection of games, we formalized their rulesets and mechanics into programmatic environments that the LLM agents could interact with.

Our final selection of games were \textit{Air, Land, Sea} (ALS); \textit{Arctic Scavengers} (ARC); \textit{Are You the Traitor?} (AYT); \textit{Codenames} (CN); \textit{Hive} (HV); \textit{Pit} (PT); \textit{Santorini} (SN); \textit{Two Rooms and a Boom} (TRB); and \textit{Sea Battle} (SB). Descriptions of the games and their rules can be found in Appendices \ref{game_descriptions} and \ref{game_rules} respectively. For additional details about game implementation, see Appendix \ref{implementation}.

\begin{table}[h]
  \caption{\textbf{Number of games per reasoning category} We identify a set of six orthogonal components of strategic reasoning and curate a set of games that sufficiently cover their spread.}
  \vspace{.5em}
  \label{game-properties-table}
  \centering
  \begin{tabular}{lcc}
    \toprule
    \textbf{Reasoning Category} & \makecell{\textbf{Total}} & \makecell{\textbf{Games}} \\
    \midrule
    Abstract Strategy        & 6 & ALS, ARC, CN, HV, SN, SB \\
    Non-Deterministic        & 3 & ARC, TRB, SB \\
    Hidden Information       & 3 & ARC, AYT, TRB \\
    Language Communication   & 4 & AYT, CN, PT, TRB \\
    Social Deduction         & 2 & AYT, TRB \\
    Cooperation              & 4 & AYT, CN, SB, TRB \\
    \bottomrule
  \end{tabular}
\end{table}

\subsection{API}
\label{api}

Each environment, implemented in Python, describes a \texttt{Game} object with methods for initializing, retrieving the game's current state and available actions, updating the state with an action, and executing a full match between two agents. Agents are objects that describe a method for choosing an action conditioned on the rules, state, and available actions retrieved from a \texttt{Game} instance. Agents are instantiated at the beginning of a match and destroyed at the end, so agents may maintain persistent state between moves to choose an action.

\subsection{Rating calculation}
\label{rating_model}

We formalize our rating calculation as follows. Let our dataset contain $P$, the population of all possible matches across all games, and $S=\{m_1, m_2, \ldots, m_n\}$, our sample of $n$ matches. Define the weight $w_i$ for each match $m_i$ to be inversely proportional to the number of matches collected for that match's game. Specifically, if match $m_i$ belongs to game $X$ which has $N_X$ matches, then the weight $w_i$ is given by:
\begin{equation}
w_i=\frac{1}{N_X}.
\end{equation}
We then perform bootstrapping on the sample $S$ for $B=10,000$ times. Let $S^*_b=[m_{i_1}, m_{i_2}, \ldots, m_{i_n}]$ be the $b$th bootstrapped sample, where $m_{i_j}$ is randomly selected from $S$ with probability proportional to $w_i$ with replacement.

\begin{equation}
    P(i>j)=\frac{e^{\beta_i}}{e^{\beta_i}+e^{\beta_j}}
\end{equation}

For each bootstrapped sample $S^*_b$, we use maximum-likelihood estimation to fit the parameters of the above exponential Bradley–Terry model $\theta_b=\{\beta_\text{random}, \beta_\text{GPT-3}, \ldots\}$. Let $\theta_{b,k}$ denote the parameter for agent $k$ in bootstrapped sample $b$. We take the means of these distributions to be the "true" rating $\hat{\theta}_k$ for each agent $k$, given by:
\begin{equation}
\hat{\theta}_k=\frac{1}{B}\sum_{b=1}^{B}\theta_{b,k}
\end{equation}

We considered several methods for aggregating pairwise match results across games into scores that represent the general skill of each model, including the Elo system \citep{elo1967proposed}. Unlike Elo, the Bradley–Terry system \citep{bradleyterry} assumes model skill does not change over time and it does not need to be calculated in a decentralized manner, making it more suitable for evaluating language models \citep{chiang2023chatbot}. In our analysis, this model also enables the comparison of models that never directly competed.

\section{Empirical results}
\label{empirical_results}

Additional figures showing agent-pairwise data covering number of games, total score, win probability, and rating per game is available in Appendix \ref{additional_figures}. The rating plots in Appendix \ref{additional_figures} show 90\% confidence intervals for the points in Figure \ref{fig:rating_scatter}.

\begin{table}[h]
\caption{\textbf{Game ratings} The table highlights the effects of scaffolds. Across all games, GPT-4 with CoT scaffolding improves over the base model substantially. But GPT-3 with CoT scaffolding is outperformed by the base model in \textit{Air, Land, and Sea}, \textit{Hive}, and \textit{Two Rooms and a Boom}. Additionally, GPT-4 with RAP scaffolding usually under-performs GPT-4-CoT except in \textit{Are You the Traitor?}, \textit{Sea Battle}, and \textit{Two Rooms and a Boom}.}
\vspace{.5em}
\centering
\begin{tabular}{lcccccccccc}
\toprule
Agent &  & \multicolumn{9}{c}{Rating} \\
\cmidrule(lr){2-11}
 &  Overall & ALS & ARC & AYT & CN & HV & PT & SN & TRB & SB \\
\midrule
random & -0.50 & 1.07 & \textbf{0.48} & -2.52 & -2.67 & -1.15 & \underline{0.63} & 0.37 & -0.79 & 0.05 \\
human & \textbf{1.76} & \underline{1.49} & \underline{0.45} & 1.92 & 1.26 & \textbf{3.63} & \textbf{1.29} & -0.89 & \underline{1.70} & \textbf{1.25} \\
gpt-3 & -0.48 & 1.26 & -0.05 & -1.84 & -2.06 & \underline{1.27} & \underline{0.63} & -0.01 & -2.51 & -0.41 \\
gpt-3-cot & 0.06 & 0.03 & 0.22 & \underline{2.42} & 0.45 & -0.44 & \underline{0.63} & \underline{0.53} & -2.76 & 0.26 \\
gpt-4 & -0.89 & -7.38 & -0.12 & -2.73 & -0.65 & -1.31 & -4.42 & -0.08 & 0.62 & -1.40 \\
gpt-4-cot & \underline{0.16} & \textbf{2.13} & 0.27 & -0.19 & \textbf{2.41} & -1.13 & 0.63 & -0.53 & 1.22 & \underline{0.62} \\
gpt-4-rap & -0.10 & 1.41 & -1.25 & \textbf{2.94} & \underline{1.26} & -0.86 & \underline{0.63} & \textbf{0.62} & \textbf{2.51} & -0.37 \\
\bottomrule
\end{tabular}

\label{tab:rating}
\end{table}

\begin{table}[h]
\caption{\textbf{Average score.} The total score an agent achieved in a game divided by the number of games that agent played. Comparing with \ref{tab:rating}, this table highlights interesting correlations between empirical score and model-inferred ratings. For example, in \textit{Air, Land, and Sea}, GPT-4-CoT has the top rating while the human baseline has second-top, but they swap when examining average score. This plot also shows more clearly why the human baseline has the highest rating even though both the human baseline and GPT-4-RAP have the highest rating in three games. Here, the human baseline achieved the highest score in four games but GPT-4-RAP only achieved the highest in two.}
\vspace{.5em}
\label{tab:score}
\centering
\begin{tabular}{lcccccccccc}
\toprule
Agent &  & \multicolumn{9}{c}{Score} \\
\cmidrule(lr){2-11}
 &  Overall & ALS & ARC & AYT & CN & HV & PT & SN & TRB & SB \\
\midrule
random & 0.49 & 0.72 & \textbf{0.60} & 0.25 & 0.18 & 0.41 & \underline{0.50} & 0.56 & 0.52 & \underline{0.58} \\
human & \textbf{0.85} & \textbf{1.00} & NaN & NaN & NaN & \textbf{1.00} & \textbf{1.00} & 0.43 & NaN & \textbf{0.78} \\
gpt-3 & 0.48 & 0.64 & 0.43 & 0.43 & 0.63 & \underline{0.80} & \underline{0.50} & 0.47 & 0.27 & 0.40 \\
gpt-3-cot & 0.60 & 0.43 & \underline{0.50} & \underline{0.93} & \underline{0.89} & 0.60 & \underline{0.50} & \textbf{0.61} & 0.33 & 0.55 \\
gpt-4 & 0.31 & 0.00 & 0.42 & 0.33 & 0.83 & 0.33 & 0.31 & 0.42 & 0.71 & 0.20 \\
gpt-4-cot & 0.60 & \underline{0.81} & \underline{0.50} & 0.64 & \textbf{1.00} & 0.50 & \underline{0.50} & 0.37 & \underline{0.75} & 0.51 \\
gpt-4-rap & \underline{0.62} & NaN & 0.33 & \textbf{1.00} & NaN & 0.50 & NaN & \underline{0.58} & \textbf{1.00} & 0.26 \\
\bottomrule
\end{tabular}

\end{table}

\subsection{Human comparison}

The human baseline outperforms all model and scaffolding configurations in the benchmark. The upper-bound of GPT-4-RAP's confidence interval in Figure \ref{fig:subim2} just reaches the lower-bound of the human baseline. But due to both GPT-4-RAP and the human baseline having very few data points, this detail should not be taken very seriously. In Table \ref{tab:score}, the human baseline achieves the highest overall score in every game it played except for \textit{Santorini}. 

The human subject beat their opponent agent in all matches except for two of the three \textit{Codenames} matches. For these particular matches, the human subject employed a friend because \textit{Codenames} typically requires at least two players per team. We hypothesize that LLM agents perform better in this context because they are better at modeling their teammate's thought process, as they are instantiated from the same underlying language model. In contrast, pairs of humans share much less cognitive similarity.

Details about the human data collection process are discussed in Appendix \ref{human}.

\subsection{Effect of scaffolding}
\label{cot_effect}

Chain-of-Thought prompting provided the best median and upper quartile results of all configurations tested in Figure \ref{fig:subim2}. GPT-3 and GPT-4 showed almost identical performance with GPT-4 with only a slight improvement over GPT-3. The positive effects of CoT prompting are already well-documented \citep{chowdhery2022palm, zelikman2022star, wei2022emergent}, and our results provides evidence of their use in strategic settings.

If we interpret the addition of CoT scaffolding as an intervention on the base model, we see it improves strategic reasoning ability in GPT-4 moreso than in GPT-3. In \textit{Sea Battle}, this intervention brings GPT-4 from the worst model to the best model. In every game except \textit{Codenames}, GPT-4 with CoT scaffolding outperforms its base model. But for GPT-3, the base model outperforms the CoT variant in \textit{Santorini} and \textit{Sea Battle}. One possible hypothesis for this difference in effect between on GPT-3 and GPT-4 is that GPT-4 is a bigger model and thus can probably make better use of in-context information.

\subsection{GPT-3 versus GPT-4}

GPT-3 performs only slightly better than random action. Surprisingly, GPT-4 performs the worst of all configurations with its upper quartile performance being worse than random's lowest quartile. This result is mostly due to GPT-4 losing all matches in Sea Battle. This challenges our aggregation method: GPT-4 should not be so harshly penalized for poor performance on one game.

An alternative aggregation method that would be more robust to outliers is to use factor analysis to isolate a "general strategic reasoning factor" that explains a significant portion of the variance between models' performances. This method is used to aggregate separate cognitive test scores into IQ scores, making it apt for evaluating LLMs' reasoning abilities \citep{ilić2023unveiling}. We expect this g-factor approach to appropriately weigh models' \texttt{Sea Battle}  ratings lower, fixing this discrepancy.

Considering these two datapoints and analysis from \ref{cot_effect}, we can tentatively conclude that strategic reasoning ability is not improving in OpenAI's newest frontier models alone, but their receptiveness to scaffolding to improve strategic reasoning is increasing.

\subsection{State-of-the-art scaffolding}

The state-of-the-art scaffolding was outperformed by both Chain-of-Thought agents. One possible hypothesis for this is that, during the Monte-Carlo tree search, this agent predicts new states based on the state being examined, which is already a predicted state depending if depth $\geq 1$. If the agent makes any errors in this examined state's prediction due to misunderstandings about the game state or rules, these will likely be compounded in the next set of predictions. We might expect the Chain-of-Thought agents to be susceptible to the same issue of compounding errors, but to a lesser extent. This could be tested qualitatively by a human expert analysing GPT-4-RAP's predictions for accuracy.

Another hypothesis is that we ran GPT-4-RAP to a depth great-enough to surpass GPT-4 without RAP scaffolding, but not great-enough depth to surpass Chain-of-Thought scaffolding. This could be tested by adding several GPT-4-RAP agents to the benchmark, each with different depths.

It seems unlikely that Chain-of-Thought prompting should be the most sophisticated black-box scaffolding, so it remains an open question to find this scaffolding in order to establish an upper-bound on strategic reasoning ability with black-box scaffolds.

\section{Discussion}
\label{limitations}

We now discuss the limitations and future directions of our work.

\textbf{Confirming out-of-distribution status} It is clear by simply asking GPT-4 that it already knows about these games and their rules. It is unclear, however, if it consumed any strategy guides about these games in the pretraining process, which is the determining factor for out-of-distribution status in our benchmark. \textbf{Future work} We propose the following experiment. Design an intervention that is: supply a strategy guide in-context to a language model agent for the game it is playing. We would expect this intervention to improve agent performance more on out-of-distribution games than in-distribution games. Collect data of agents playing an unknown distribution game; agents with the intervention playing an unknown distribution game; agents playing known in-distribution games; agents with the intervention playing known in-distribution games. Compare the effect of the intervention on the unknown distribution game versus the effect on the known in-distribution games. If the effect is much higher on the unknown distribution game, this is a evidence for the game being out-of-distribution. This would work better with known out-of-distribution games, but this may not be possible to know in all cases. We could also compare models' performance on common games vs. "counterfactual" games, which are slightly modified to reduce any association with their in-distribution counterparts \citep{Wu2023ReasoningOR}.

\textbf{Protecting out-of-distribution status} We did not attempt to protect these games from becoming in-distribution in the future. \textbf{Future work} Developers of frontier models should curate strategic reasoning environments by ensuring these games are held out from pretraining data. For ubiquitous games such as chess, this is unfeasible. But following our heuristics for game selection discussed in section \ref{game_selection}, it should be reasonable to find games without much internet data.

\textbf{Results' sensitivity to games} From inspecting GPT-4's surprisingly low rating with \textit{Sea Battle}, it became apparent that our "multigame" approach to aggregation may be inadequate due its sensitivity to the games included; i.e., ablating \textit{Sea Battle} significantly changed the data narrative. \textbf{Future work} We see multiple ways forward. If aggregate data is useful, investigate more robust forms of aggregation, such as the g-factor or factor analysis in general. Alternatively, explore a multi-dimensional approach that attempts to score agents on the six reasoning categories from Table \ref{game-properties-table}. Or, discard any notion of aggregation and determine effective means of analysis that looks only at individual games and maybe uses more qualitative data with the help of human experts.

\textbf{Low-resolution human benchmark} We find it especially important to know how well these models fair compared to humans, but collecting comprehensive human data was out of our means. \textbf{Future work} Conduct more comprehensive human data to form a distribution of human strengths on each game with which we can measure the progress of model and scaffolding development.

\textbf{Uncaught edge cases} Every few games were inspected during data collection, and occasionally, we caught and fixed bugs in our evaluation code. It is possible that some edge cases went unnoticed and were featured in our final data release. \textbf{Future work} Incorporating more human subjects into the data collection should make this process trivial, as they can provide immediate feedback if they witness unexpected behavior.

\textbf{Benchmark and dataset size} Our benchmark has a respectable number of games and agents compared to other benchmarks \citep{Chen2024LLMArenaAC, Duan2024GTBenchUT, Abdulhai2023LMRLGB, Xu2023MAgICIO}, but the addition of more games and agents would provide a richer picture of models' strategic reasoning abilities. Additionally, our dataset is fairly small and suffers from biases from variable resource cost between games. \textbf{Future work} Add more varied games to the benchmark, evaluate more model and scaffolding configurations, and collect more data for each configuration.

\section{Conclusion}
\label{Conclusion}

We present \textsc{GameBench}, an LLM agent benchmark to test strategic reasoning ability using diverse games that have sparse strategy material in pretraining data. We benchmark OpenAI’s GPT-3 and GPT-4 models and evaluate the impact of two scaffolding methods: Chain of Thought (CoT) and Reasoning via Planning (RAP). We find that human trials consistently outperform all LLM agents. Of all the agent configurations, CoT agents performed the best, followed by RAP-augmented GPT-4. Base GPT-3 performed on-par with the random baseline, and base GPT-4 performed worse. These results show that while measures such as scaffolding can help improve performance in strategic reasoning, even the best configuration fall short of human reasoning. LLMs show great promise working on in-distribution tasks, though their performance on OOD task sets show a low risk for current dangers of deploying autonomous agents. Nonetheless, the performance gains achieved through scaffolding techniques indicate the potential for future advancements that could increase the risk posed by such systems if their reasoning capabilities continue to improve.

\begin{ack}

We thank Misha Gerovitch and Severin Field for providing feedback on our drafts. We thank Shubhorup Biswas for implementing Atari Boxing.
\end{ack}



\bibliography{citations} 

\appendix

\section{Hazards}
\label{hazards}

We believe that good strategic reasoning is a dangerous capability for an AI agent to have, especially one that will operate autonomously. Thus, good performance on this benchmark could correlate with harmful risk. This is important for labs developing frontier models to be able to measure and be aware of, but it is also possible for a malicious or ignorant actor to use this benchmark as a feedback signal to improve their own large language model's strategic reasoning ability. However, we think that the former benefit outweighs the latter risk in this time where the development of large language models is largely controlled by a few frontier labs. And we reduce the risk from ignorant actors by producing these benchmarks and discussing their importance.

\section{Research on human subjects}
\label{human}

Our human-based data-points came from a co-creator of the benchmark, and the same person with their friend for \textit{Codenames}.

The instructions were communicated informally because the subject co-designed the benchmark and this human study. They were initially instructed to play against the GPT-4-RAP, but due to resource costs, were later instructed to play against the any agent except GPT-4-RAP or the random baseline. They were instructed to not play \textit{Are You the Traitor?} and \textit{Two Rooms and a Boom} because they are social deduction games and it is not a good setup to have one agent with extra information than other agents. They were instructed to collect as many matches as they were willing to collect in the time they had available.

No additional compensation was provided to them for data collection, but the API costs were covered.

Given the informal nature of the data collection, the near-zero risk, and the fact that the subject was a co-creator in this benchmark and this experiment, we did not discuss risks or consult an IRB. The data this person created do not contain any identifying information.

\section{Dataset documentation}
\label{data_doc}

The data used to generate the figures and tables in this paper are available in our Github \url{https://github.com/Joshuaclymer/GameBench} under the CC-BY 4.0 license. These data will remain available here as long as Github is available. New data may be added by the authors in the future, which will be documented in the commits.

The intended use of this data is to compare GPT-3 and GPT-4 on this benchmark, and to compare against new models, scaffolds, baselines, and informed-and-consenting humans in the future.

The data are in JSON format. The top-level object is an array, and the array contains objects. Each object has a \texttt{"game"} key which indicates the game, and two other keys -- the two agents that played in no particular order -- with their respective score as the value. Scores are in the range \texttt{[0, 1]} and sum to 1.

Our data collection was not uniform across games nor against agent-pairs due to resource constraints. In general, we preferred playing agents against the random baseline and preferred games that didn't take too long to complete.

All data for each agent except the random agent were collected using OpenAI's completions API. Each game was designed to take no more than ~5 minutes when playing base GPT-4 against random. Cost estimates were not obtained, but it can be assumed that CoT agents will cost approximately twice their base variants, and GPT-4-RAP will cost approximately \texttt{base cost x MCTS depth x number of actions per state x 6}

\begin{figure}[h]

\begin{subfigure}{0.5\textwidth}
\includegraphics[width=1\linewidth, height=6cm]{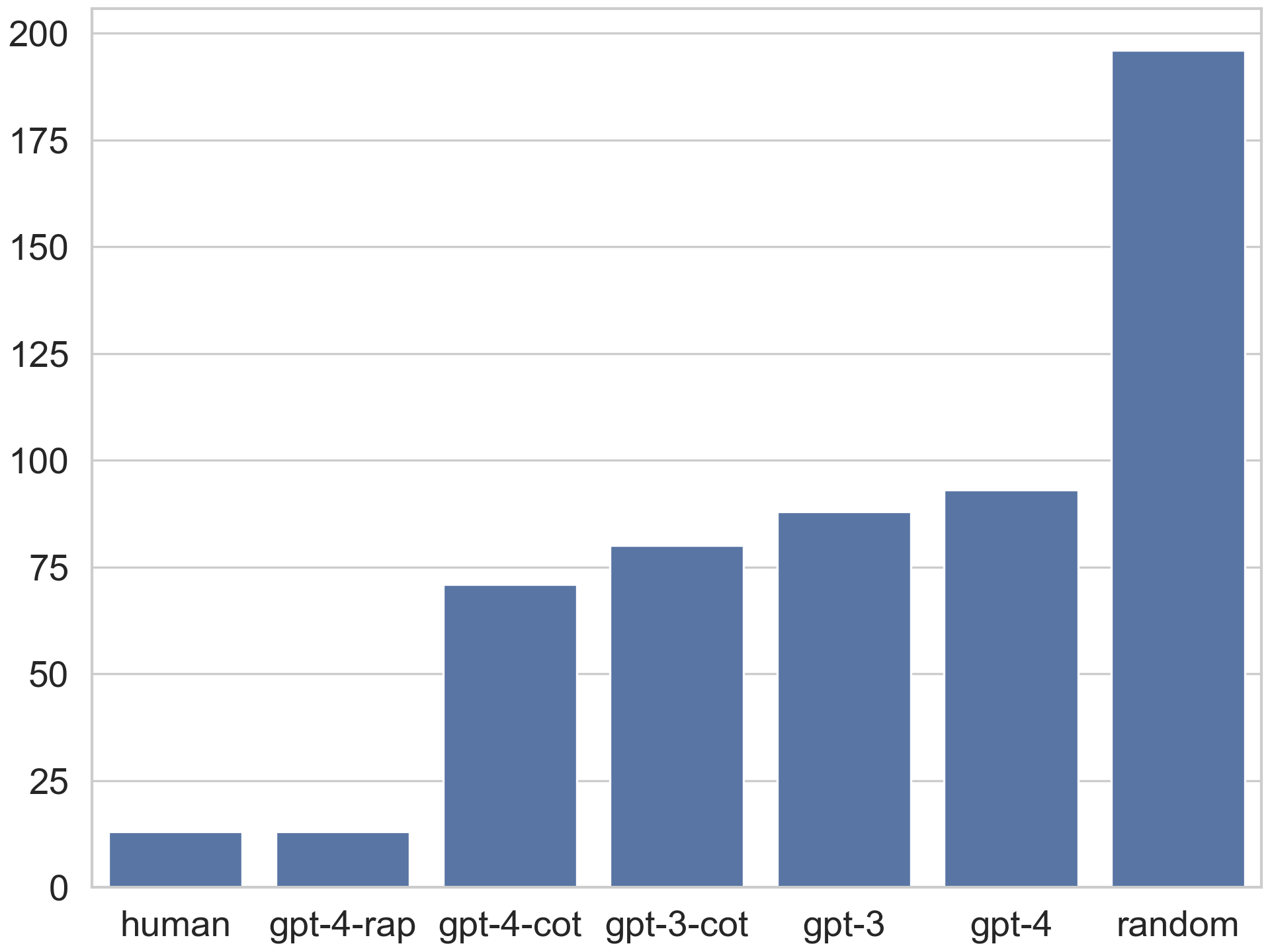} 
\caption{Per agent}
\end{subfigure}
\begin{subfigure}{0.5\textwidth}
\includegraphics[width=1\linewidth, height=6cm]{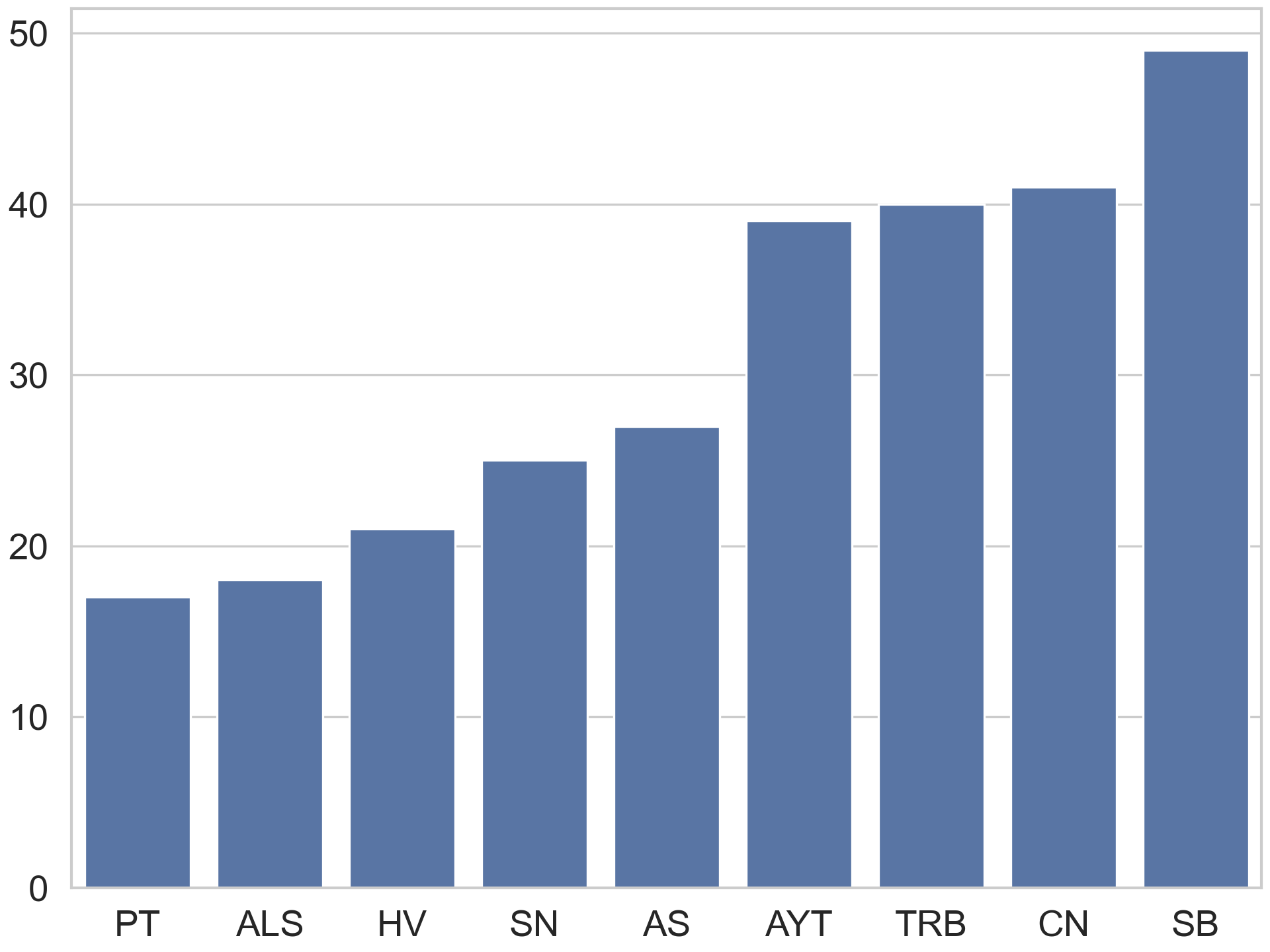}
\caption{Per game}
\end{subfigure}

\caption{\textbf{Number of matches recorded} The random baseline and faster games were oversampled due to their low cost.}
\end{figure}

\section{Additional implementation details}
\label{implementation}

To measure multimodal capabilities, \textit{Hive} was made to use images to represent its game state, instead of text like all the other games. However, GPT-3 is not multimodal, so it was served textual representations of the graphical state created by GPT-4. Then, for RAP, GPT-4 with the completions API can't produce images when predicting future states, so for simplicity, the image is turned into a text description here as well.

GPT-4-RAP was run with the default parameters from the \texttt{llm-reasoners} library \citep{maitrix-org_llm-reasoners_2024} except the Monte-Carlo tree search depth limit was set to 2 due to resource constraints.

Data was not collected for a GPT-3-RAP because GPT-3 refused to comply with prompts asking it to predict actions, game states, or other players' behaviors. The model would often reply, "As a language model, I can incapable of predicting..." Because it is unlikely that GPT-3 is self-aware and because we never indicate to the model that it is a language model in our prompting, we hypothesize that this refusal is due to the nature of GPT-3's hidden system prompting for refusing unsafe behaviors and not due to a lack of ability on GPT-3's part. As such, it is difficult to measure the relative effect of RAP scaffolding on GPT-4 versus GPT-3.

CoT-scaffolded agents are prompted with \texttt{"First, let's reason out loud about which action you should take to maximize your probability of winning."} after they see the game state and available actions.

GPT-4-RAP employs a Monte-Carlo tree search where states and actions are model predictions, and rewards are computed using next-token probabilities. Our implementation of RAP relies heavily on code from \citet{hao2023reasoning}. Their code is available under the Apache License 2.0. Details of our prompting strategy with RAP can be found in appendix \ref{rap_prompt}.

We use the Python library \texttt{choix} \citep{choix} to find the maximum-likelihood estimate of agent ratings in the Bradley–Terry model. This library is made available under the MIT License.

There is one extra game in the benchmark that can be found on the Github repository that was not included in data collection: \textit{Atari Boxing}. Data collection on this game turned out to be too cumbersome, but as the only real-time game, it measures a factor not covered by the other games, and thus is important for future benchmarking.

All games received two agents regardless of team size. In cases with multiple cooperative players on one team, the agent is duplicated. The agent is not made explicitly aware that it is duplicated.

All code for running the benchmark on existing models and scaffolds, for creating implementing new agents, and for reproducing results can be found in our Github \url{https://github.com/Joshuaclymer/GameBench}

\section{RAP prompting}
\label{rap_prompt}

Reasoning-via-Planning describes a framework for using a probabilistic language model in a Monte Carlo tree search. Exactly how the model is prompted depends on the implementation. Below, the [rules] and [rules subtopics] come from Appendix \ref{game_rules}.

\textbf{Rules subtopics}

\fbox{\begin{minipage}{\dimexpr\linewidth-2\fboxsep-2\fboxrule\relax}
If you would like to learn more about the rules at any point, use rule(<topic>) where <topic> is one of [subtopics from rules above].
\end{minipage}}

\textbf{\texttt{!PREFIX}}

\fbox{\begin{minipage}{\dimexpr\linewidth-2\fboxsep-2\fboxrule\relax}
You are now playing a game called [title]. The rules are as follows [rules summary from game above]. [Rules subtopics]. Your observation of the game is between <state> and </state>: <state>[game state]</state>
\end{minipage}}

\textbf{\texttt{!EXAMPLE}}

\fbox{\begin{minipage}{\dimexpr\linewidth-2\fboxsep-2\fboxrule\relax}
You are playing a game called monty hall. The rules of the game are as follows: there are three doors, behind one of which there is a prize. Select the door with the prize. Your observation of the state is between <state> and </state>: <state>All three doors are closed.</state>
\end{minipage}}

\textbf{Prompt for list of available actions}

\fbox{\begin{minipage}{\dimexpr\linewidth-2\fboxsep-2\fboxrule\relax}
\texttt{!EXAMPLE}

\texttt{User:} To the best of your ability, predict the available actions in this position between <action> and </action>:

\texttt{Assistant:} <actions> 0. Choose the left door 1. Choose the middle door 2. Choose the right door</actions>

\texttt{!PREFIX}

\texttt{User:} To the best of your ability, predict your available actions in this position between <actions> and </actions>:
\end{minipage}}

\textbf{Prompt for selecting an action} The next-token probability from this prompt is used to in the reward calculation.

\fbox{\begin{minipage}{\dimexpr\linewidth-2\fboxsep-2\fboxrule\relax}

\texttt{!PREFIX}

\texttt{User:} To the best of your ability, predict your available actions in this position between <actions> and </action>:

\texttt{Assistant:} <actions>[actions from previous model prediction, enumerated]</actions>

\texttt{User:} Choose an action by writing only the associated number.

\end{minipage}}

\textbf{Prompt for self-evaluating an action} The next-token probability from this prompt is used in the reward calculation.

\fbox{\begin{minipage}{\dimexpr\linewidth-2\fboxsep-2\fboxrule\relax}
\texttt{User:} Write your action below:

\texttt{Assistant:} [action from previous model prediction]

\texttt{User:} Is this a good action? yes/no.
\end{minipage}}

\textbf{Prompt for guessing other players' actions}

\fbox{\begin{minipage}{\dimexpr\linewidth-2\fboxsep-2\fboxrule\relax}
\texttt{!EXAMPLE}

\texttt{User:} To the best of your ability, predict what actions other players might take between <others> and </others>:

\texttt{Assistant:} <others>My opponent is going to reveal one of the two doors I don't choose.</others>

\texttt{!PREFIX}

\texttt{User:} To the best of your ability, predict what actions other players might take between <others> and </others>:
\end{minipage}}

\textbf{Prompt for guessing the game state}

\fbox{\begin{minipage}{\dimexpr\linewidth-2\fboxsep-2\fboxrule\relax}
\texttt{!EXAMPLE}

\texttt{User:} Write your action below:

\texttt{Assistant:} I will choose the left door

\texttt{User:} Write other player's actions below:

\texttt{Assistant:} My opponent will reveal the middle door

\texttt{User:} To the best of your ability, predict your new observation of the game based on your actions and others' actions between <state> and </state>:

\texttt{Assistant:} <state>\escape{n}The left and right doors are closed, and the middle is open. There is no prize behind it.\escape{n}</state> 

\texttt{!PREFIX}

\texttt{User:} Write your action below:

\texttt{Assistant:} [action from previous model prediction]

\texttt{User:} Write other players's actions below:

\texttt{Assistant:} [other players' actions from previous model predictions]

\texttt{User:} To the best of your ability, predict your new observation of the game based on your actions and others' actions between <state> and </state>:
\end{minipage}}

\textbf{Prompt for open-ended actions} The API we designed allows games to give "open-ended actions" to agents, in which they don't select from a predefined list of options but instead provide a text response as the action. However, RAP doesn't support this format, so we convert open-ended actions into ones with predefined options by prompting the model for a response to the open-ended action before feeding it into the Monte Carlo tree search algorithm.

\fbox{\begin{minipage}{\dimexpr\linewidth-2\fboxsep-2\fboxrule\relax}
\texttt{!EXAMPLE}

\texttt{User:} Write your action below:

\texttt{Assistant:} Ask my opponent a question.

\texttt{User:} This is an openended action. Write a description of what you're going to do.

\texttt{Assistant:} I will pretty-please ask them to tell me which door has the prize.

\texttt{!PREFIX}

\texttt{User:} Write your action below:

\texttt{Assistant:} [openended action from game]

\texttt{User:} This is an open-ended action. Write a description of what you're going to do.
\end{minipage}}

\textbf{Prompt for assessing win probability} The next-token probability from this prompt is used in the reward calculation.

\fbox{\begin{minipage}{\dimexpr\linewidth-2\fboxsep-2\fboxrule\relax}
\texttt{!PREFIX}

\texttt{User:} Will you eventually win from this position? yes/no
\end{minipage}}

\section{Game descriptions}
\label{game_descriptions}

\textbf{Air, Land, and Sea} is a war strategy game where players are Supreme Commanders fighting to control two of three areas (air, land, sea) by deploying limited Battle card forces each round. The first commander to accumulate 12 points across multiple battles wins the war. \url{https://boardgamegeek.com/boardgame/247367/air-land-and-sea}

\textbf{Arctic Scavengers} is a resource-management game in which players are the leader of a small tribe of survivors. Resources, tools, medicine, and mercenaries are all in scarce supply. Players are pitted against each other in a fight for survival. The agent with the largest tribe at the end of the game is declared the winner and receives 1 point. \url{https://www.riograndegames.com/games/arctic-scavengers-with-recon-expansion/}

\textbf{Are You the Traitor} is a social deduction game where players are secretly divided into Good and Evil teams. The players then engage in an unstructured conversation trying to deduce the opposing team's critical roles. A player will yell "Stop!" while pointing at someone, and that round ends. If they identify their target role correctly, their team earns Treasure cards. The team with the most Treasure after multiple rounds wins. \url{https://www.looneylabs.com/games/are-you-traitor}


\textbf{Codenames} is a 2v2 cooperative game with one spymaster and one operative per team. All players see a grid of words, and it is the spymasters’ job to create one-word clues that relate to multiple predetermined words from the grid at once, and operatives must keep using these clues to guess all of their team’s words. Agents are awarded more points for correctly guessing more words. \url{https://boardgamegeek.com/boardgame/178900/codenames}

\textbf{Hive} is a strategy game occurring on a hexagonal grid. Each player has a team of bugs, each with a unique skillset. Players try to coordinate their bugs in order to completely surround the enemy’s queen bee. The winning agent is awarded 1 point. \url{https://www.gen42.com/games/hive}

\textbf{Pit} is an every-person-for-themselves trading simulation. Each player has a hand of cards, and each card represents a certain commodity in the market. Players must trade semi-blindly trade cards to try to obtain enough of any commodity to “corner the market.” Agents are awarded points based on the commodity that they corner the market with. \url{https://www.gamenightgames.com/win1012.html}

\textbf{Santorini} is a strategy game in which two players take turns moving one of their two pawns on a five by five grid and building blocks on the grid. The game ends when one of the players moves a pawn to a square that has been built three blocks high or when one of the players cannot make a move. The winning agent is awarded 1 point. \url{https://roxley.com/products/santorini}

\textbf{Two Rooms and a Boom} is a cooperative social-deduction game in which all players are split into two teams and then mixed around between two rooms. No two players start knowing other players’ teams or roles on the team, but it is the red team’s goal to end the game with the red-team bomber and blue-team president in the same room, and it is the blue team’s goal for the opposite. The winning agent is awarded 1 point for satisfying their team’s objective. \url{https://www.tuesdayknightgames.com/products/two-rooms-and-a-boom}

\textbf{Sea Battle} is 3v3 board game in which players’ attempt to sink their opponents’ ships and their movement and cannon-firing actions occur simultaneously. The winning agent is awarded 1 point if they eliminate all enemy ships before themselves becoming eliminated. \url{https://yppedia.puzzlepirates.com/Sea_battle}

\section{Game rules}
\label{game_rules}
The rules as follows are exactly as they were shown to the language models. Rules in bullet points were withheld until requested by a model taking a specific action "Explain(\textit{rule heading})".

\textbf{Arctic Scavengers} 
        The game is played in 6 rounds, with each round consisting of a resource gathering phase and a skirmish phase.
        In the resource gathering phase, players draw cards from their deck and take actions to gather resources from the mercenary piles and the junkyard pile.
        In the skirmish phase, players compare the strength of their tribes and the winner of the skirmish gains a contested resource card.
        The game ends when all contested resource cards have been won, and the player with the largest tribe is the winner.

\textbf{Are you the traitor?} The Good team wants to destroy an Evil Magic Key while the Evil team wants to keep it. The key can be destroyed by giving it to the Good Wizard, but there is an Evil Wizard who looks exactly alike. Use social deduction to find out who is who, but also know that there is a traitor among the guards who have the key.

\textbf{Two Rooms and a Boom} 
        Two teams, Blue and Red, have opposing goals. At the end of three rounds the Red team wants to have both the President and the Bomber in the same room, while Blue team wants them to be in opposite rooms.
        Each round will allow the Leader of each room to trade 'hostages' in order to find out who the President and Bomber are and use that info to achieve their team's mission.
        Find out information by talking to other hostages in your room.

\textbf{Air Land and Sea} A strategic card game where two players compete over a series of battles to control different Theaters of war: Air, Land, and Sea. Each player is dealt 6 cards representing various military units and tactics. Players win a battle by controlling more Theaters than their opponent or convincing their opponent to withdraw. Victory Points (VPs) are earned by winning battles, and the first player to reach 12 VPs wins the game. Players must carefully manage their hand and strategically deploy cards to outmaneuver their opponent.

\begin{itemize}
	\item \textbf{Battle Structure} During a Battle, the players take turns playing one card at a time, trying to control more Theaters than their opponent.You don't draw cards during a Battle, so be sure to plan carefully and make the most of the 6 cards you are dealt!
	\item \textbf{Theaters} Each of the three Theater boards creates a 'column' between the players: one for Air, one for Land, and one for Sea. These columns are called Theaters. Cards are always played into these three Theaters. If a card is in a particular Theater's column, we say that the card is 'in that Theater.'
Theaters that are next to each other are called 'adjacent Theaters.'A player owns all of the cards on their side of the Theater boards. During your turn, you will play cards only on your side of the Theaters.
	\item \textbf{Battle Cards} Cards are played to advance your war effort and how they are played will ultimately determine who wins the war (the game).
Strength: Each card has a Strength value. If the total Strength of all the cards on your side of the Theater is higher than the total Strength of all the cards on your opponent's side of that Theater, you 'control' that Theater.
Tactical Abilities: Most cards have a Tactical Ability along with Strength, which takes effect as soon as the card is played 'face up' to a Theater. These abilities are either 'Instant' or 'Ongoing.'
	\item \textbf{Type of Battle Cards} There are three types of cards: 'Air,' 'Land,' and 'Sea' cards, which relate to the three Theaters. Normally, you may only play a card 'face up' to its matching Theater: Air cards in the Air Theater, and so on.
	\item \textbf{Facedown Cards} Cards can also be played 'facedown' as a 'wild card' in any Theater. Facedown cards always have a Strength of 2. 'Facedown' cards do not have any Tactical Abilities. You may see your own facedown cards at any time, but you may not see your opponent's 'facedown' cards.
	\item \textbf{Covered Cards} When a card is played to a Theater that already contains cards, the newly played card is placed so that it overlaps the previously played card, while still showing the top portion of it. Any card overlapped by another is called a 'covered card.' Similarly, any card that is not overlapped by another card is referred to as 'uncovered.'
	\item \textbf{Resolving Battle} During a Battle, players take turns starting with the player who has the 1st Player me Commander card.
On your turn, you must take only one of these three actions: Deploy, Improvise, Withdraw.
Once you have finished your action, your opponent begins their turn. The players continue to alternate taking turns until one of them withdraws or both players have played all of their cards.
	\item \textbf{Possible actions:} Deploy: Play one card from your hand, 'face up.' When you play a card, you must follow these deployment restrictions: You can only play cards on your side of the Theater boards. The card must be the same type as the Theater you play it to. If you have other cards in that Theater already, you must place the new card so that it covers (partially overlaps) those cards.
Improvise: Play one card from your hand, 'facedown', to any Theater. 'Facedown' cards are treated as 'wild cards' and can be played to any Theater regardless of which type they are.
Withdraw: If you think your chances of winning the current Battle are low, you may withdraw. If you do, your opponent wins the Battle and gains VPs depending on how many cards are left in your hand. See the me Commander cards for more specific information.
	\item \textbf{me Commander Cards} Supreme Commander Cards: The 1st Player Supreme Commander wins tied Theaters and gains the following number of VPs based on the number of cards left in their opponent's hand if their opponent withdraws: 5+ cards = 2 VPs, 3-4 cards = 3 VPs, 2 cards = 4 VPs, 0-1 cards = 6 VPs.
The 2nd Player me Commander loses tied Theaters and gains the following number of VPs based on the number of cards left in their opponent's hand if their opponent withdraws: 4+ cards = 2 VPs, 2-3 cards = 3 VPs, 1 card = 4 VPs, 0 cards = 6 VPs.
	\item \textbf{Tactical Abilities} Most cards have Tactical Abilities described on the card. When you play a card face up from your hand, or if a facedown card is flipped over, its Tactical Ability takes effect immediately. There are two kinds of Tactical Abilities: 'Instant' and 'Ongoing', indicated on the card.
You must carry out the effects of a Tactical Ability unless they contain the word 'may'.
If a Tactical Ability is impossible to perform, that ability is ignored and has no effect.
	\item \textbf{Instant Abilities} Instant Abilities take effect immediately after the card is played or if the card is revealed by being flipped face up. Once the Instant Ability is resolved, it has no further effect (unless somehow that card is played or revealed again).
Note: Because instant abilities take effect when flipped face up, it is possible for multiple instant abilities to take effect around the same time. In these situations, always resolve the instant abilities in the order they happened and fully resolve each ability before moving on to the next.
Once an instant ability begins taking effect, it always resolves fully, even if it gets flipped facedown before completing.
	\item \textbf{Ongoing Abilities} These are always in effect as long as the card is face up. If a card with an Ongoing Ability is flipped 'facedown', the ability no longer has any effect (unless that card is revealed again).
Example: The Escalation Tactical Ability increases the Strength of all of your facedown cards to 4 as long as the Escalation card remains 'face up'. If that card were flipped over by another Tactical Ability, your 'facedown' cards would go back to being Strength 2.
	\item \textbf{Tactical Ability Key Terms} Flip: Many Tactical Abilities allow you to flip a card. Flipping a card means either turning it 'face up' if it is 'facedown' or turning a 'facedown' card so it is 'face up.'Unless the ability states otherwise, you may flip any card; yours or your opponent's.
Uncovered/Covered: Many Tactical Abilities only affect uncovered or covered cards. If an ability does not specify uncovered or covered, such as Transport or Redeploy, assume the ability can affect any card.
Play: Some Tactical Abilities instruct you to play a card, or only take effect in response to a card being played. The word 'play' describes any time a player takes a card from their hand and places it in a Theater.
Non-Matching Theaters: Means that a card is not in the Theater of its type. The card does not suffer any penalty for being in the 'wrong' Theater.
Destroy: Some Tactical Abilities instruct you to destroy a card. Destroyed cards are always placed facedown on the bottom of the deck. If a card is destroyed immediately after it is played, such as by Blockade, then that card does not get to use its Tactical Ability.
Occupied: When counting the number of cards that occupy a Theater, always count both players' cards towards that total.
Move: When a card is moved to a different Theater. It stays on the same side of the Theaters it was already on and remains owned by the same player. Moved cards are placed on top of any cards already in the Theater it was moved to. It covers those cards.
	\item \textbf{Ending Battles} There are two ways a Battle can end: If a player withdraws, their opponent wins the Battle. Or if both players have played all of the cards in their hand. At this point, the player who controls the most Theaters wins the Battle.In order to control a Theater, you must have a higher total Strength there than your opponent has in that Theater. If your Strengths are tied, the 1st Player wins the tie and controls that Theater. If there are no cards on either side of the Theater, the 1st player controls that Theater.
	\item \textbf{Scoring Victory Points} If neither player withdraws, the winner of the Battle scores 6 VPs. If one of the players withdraws, the other player scores VPs based on the number of cards left in the withdrawing player's hand (see the me Commander Cards for details). After scoring VPs, check if the victor has enough VPs to win the game (12 VPs). If they don't, fight another Battle.
	\item \textbf{Setting up Battles} All cards are collected and shuffled together to create a new deck. Deal each player a new hand of 6 cards. Next, the Theater cards are rotated clockwise so that the rightmost Theater is moved to the far left of the Theater lineup. Lastly, players swap me Commander cards. The player who was 1st in the last battle is now 2nd.
\end{itemize}
\textbf{Codenames} A strategic game of guessing and deduction where two teams, Red and Blue, compete to identify their team's words on a grid based on one-word clues given by their Spymasters. The game ends when all words of one team are guessed, or the assassin word is chosen.

\begin{itemize}
	\item \textbf{Roles} Spymaster: Knows which words correspond to which team / the assassin. Gives one-word clues that relate to any number of their team's words on the board.
Operative: Guesses words belonging to their team based on the Spymaster's clues. Aims to avoid words not belonging to their team and the assassin word.
	\item \textbf{Turn Structure} Spymaster's Turn: Give a clue to their operative and a number indicating how many words relate to that clue. Operative's Turn: Guess words, aiming to find all their team's words. After each guess, if the word is not their team's, the turn ends. If the word is their team's, they can guess again. If the word is the assassin word, the game ends and their team loses. An operative can make up to N+1 guesses, where N is the number of cards given by the Spymaster.
	\item \textbf{Winning Conditions} A team wins by correctly guessioutpg all their words. Game ends immediately if the assassin word is guessed and the team who guessed it loses.
	\item \textbf{Forbidden Actions} Spymasters cannot use part or any form of the words on the board in their clues. Spymasters cannot use words that sound like words on the board in their clues. Clues must be exactly one word and one number.
	\item \textbf{Scoring} Points are awarded based on the number of correct guesses by each team. If a team guesses the assassin word, they receive a score of 0.
	\item \textbf{Special Rules} If zero words are related to the clue, the Spymaster can give a clue of '0' and the Operative can guess an unlimited number of words.
\end{itemize}
\textbf{Hive} Hive is a bug-themed abstract strategy game. The object of Hive is to capture the opponent's queen bee by allowing it to become completely surrounded by pieces belonging to either player, while avoiding the capture of one's own queen. Tiles can be moved to other positions after being placed according to various rules, much like chess pieces.

\begin{itemize}
	\item \textbf{Placing the Queen Bee} Players must place their Queen Bee by their fourth turn. Until then, they cannot move any placed pieces.
	\item \textbf{Queen Bee Movement} The Queen Bee can only move one space at a time around the hive.
	\item \textbf{Spider Movement} The Spider can move exactly three spaces.
	\item \textbf{Ant Movement} Able to move to any empty space around the hive as long as other movement rules are not violated.
	\item \textbf{Grasshopper Movement} The Grasshopper can jump over over adjacent pieces, landing on the first empty space.
	\item \textbf{One Hive Rule} The tiles must always be connected; you cannot move a piece if it would break the hive into separate groups.
	\item \textbf{Freedom to Move} A piece can only move if it can physically slide to its new position without disturbing other tiles.
	\item \textbf{Max Turns} The game ends after 250 turns.If no Queen Bee is surrounded by the end of the game, the game is a draw.
\end{itemize}
\textbf{Santorini} Win by moving one of your pawns to the third level of the board or forcing your opponent to be unable to finish their turn. The game is played on a five by five grid of squares, and each player controls two pawns. Play alternates between the players, starting with player 1. The pawn that a player plays with alternates during each of their turns: for example, player 1 plays pawn A on their first turn, pawn B on their next turn, then pawn A, and so on. Blocks can be placed on squares on the board up to four blocks high, creating four possible height levels.

The board begins with no blocks placed, so every square begins at level 0. Before the game starts, each of the players takes turns placing each of their pawns on the board. A square is occupied if a pawn is on it.

Each turn consists of two stages: the "move" stage and the "build" stage. During the move stage, the player moves their pawn by one square (horizontally, vertically, or diagonally). They cannot move their pawn onto a square that is occupied by another pawn, more than one level higher than the pawn, or at level 4. They can move a pawn any number of levels down, to the same level, or one level higher, but not more than one level higher and not to level 4.

During the build stage, the player must select an unoccupied square adjacent to the pawn they moved during the move stage and place a block on it. They can place a block onto an unoccupied square at any level less than 4. Once a square has been built to level 4, it is "complete", meaning pawns cannot move to it and blocks cannot be placed on it. The player instantly wins if they move their pawn onto a square at level 3 or if they force their opponent to not be able to finish their turn.

\textbf{Pit} Pit is a commodity trading game where players engage in trading to accumulate points and emerge as the winner.
        The game involves commodity cards representing various goods, with each card holding a specific point value.
        Players shout out their trade offers, attempting to negotiate deals with others to acquire valuable commodities.
        Additionally, Bull and Bear cards periodically influence the market conditions, either boosting or decreasing commodity values.
        The game continues with trading phases, market fluctuations, and scoring until a player or team reaches the agreed-upon point total,
        declaring them the victor in the spirited world of commodity trading.

\textbf{Sea Battle} Sink all of your opponent team's ships before they sink all of your team's ships.

\begin{itemize}
	\item \textbf{Damage} Players can be damaged in three ways: (1) by getting shot at by another player, (2) by sailing into a rock, (3) by colliding with another ship.
	\item \textbf{Sinking} After a player has sustained enough damage, they sink and cannot play the rest of the round.
	\item \textbf{Winning} A team wins if they have at least one live ship when all of their opponents have sunken.
	\item \textbf{Board} The board is a 24x24 grid. Some squares are occupied by rocks and some are occupied by players' ships.
	\item \textbf{Gameplay} Each turn, all players choose how they want to move and how they want to shoot. All players' choices are executed simultaneously.
	\item \textbf{Teams} At the start of the game, there are three players on each team.
\end{itemize}

\section{Additional figures}
\label{additional_figures}
We present the match outcomes per game, including the number of matches, total score, win probabilities and rating per agent.

\begin{figure}
    \centering
    \includegraphics[width=0.9\textwidth]{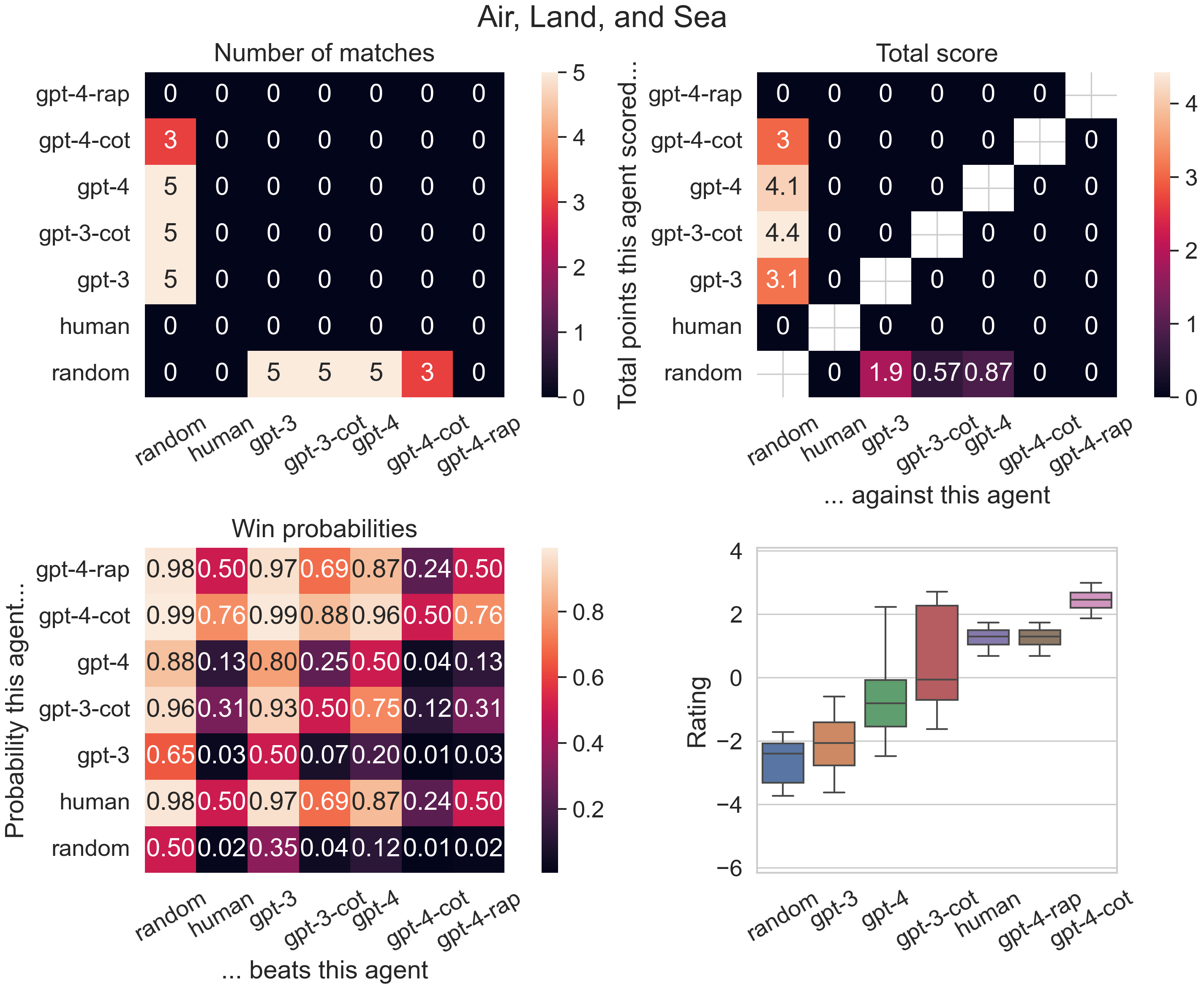}
\end{figure}
\begin{figure}
    \centering
    \includegraphics[width=0.9\textwidth]{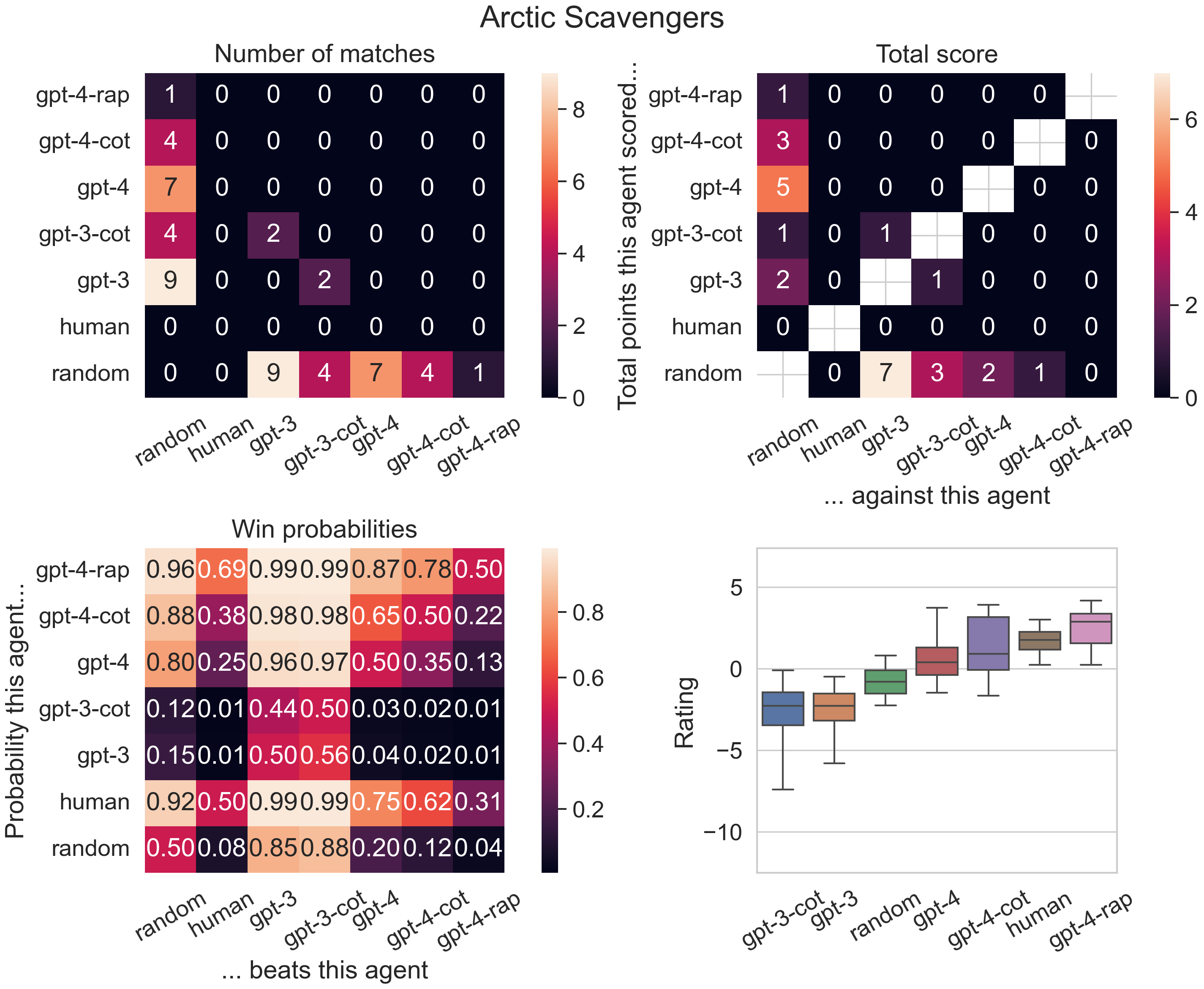}
\end{figure}
\begin{figure}
    \centering
    \includegraphics[width=0.9\textwidth]{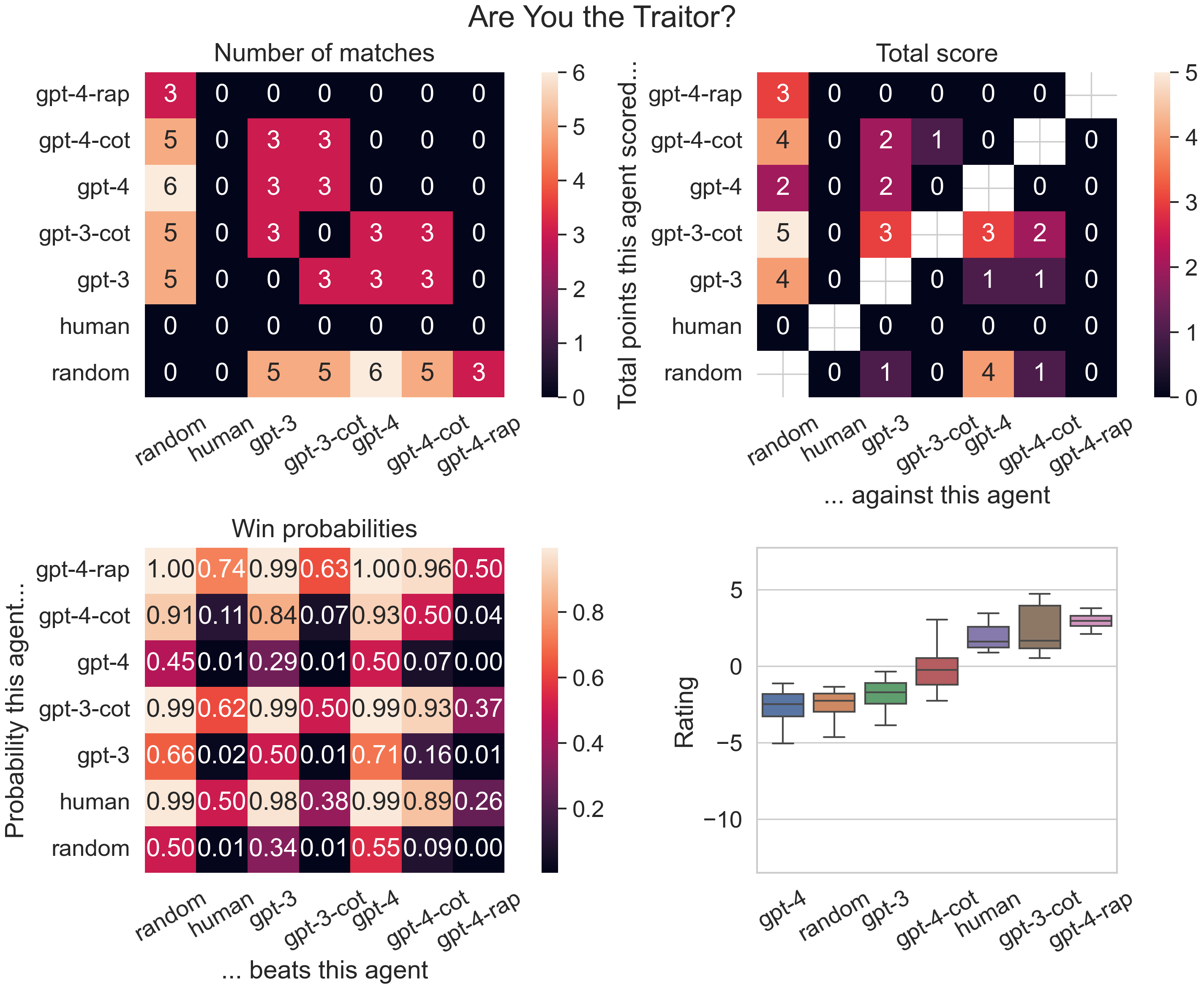}
\end{figure}
\begin{figure}
    \centering
    \includegraphics[width=0.9\textwidth]{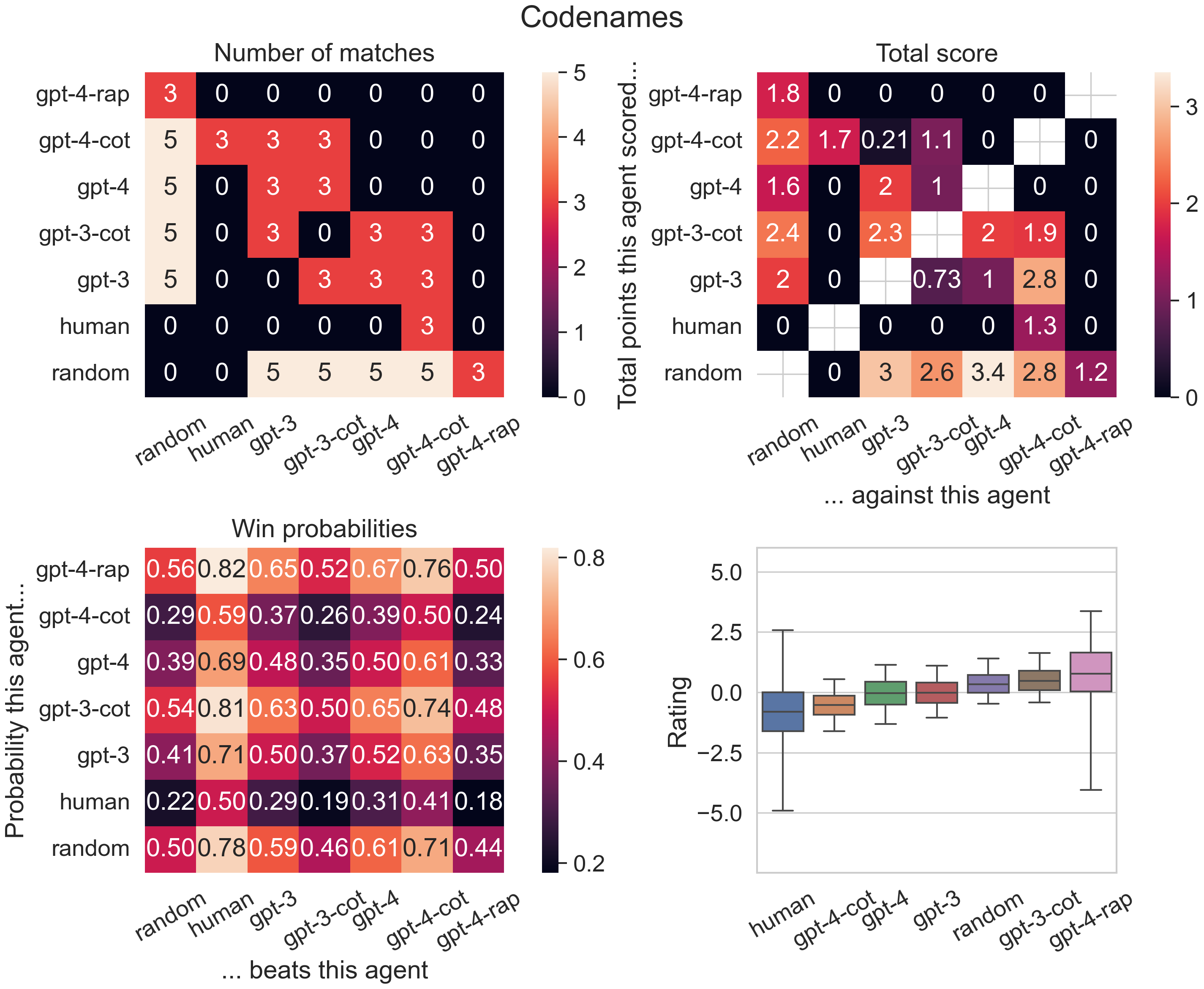}
\end{figure}
\begin{figure}
    \centering
    \includegraphics[width=0.9\textwidth]{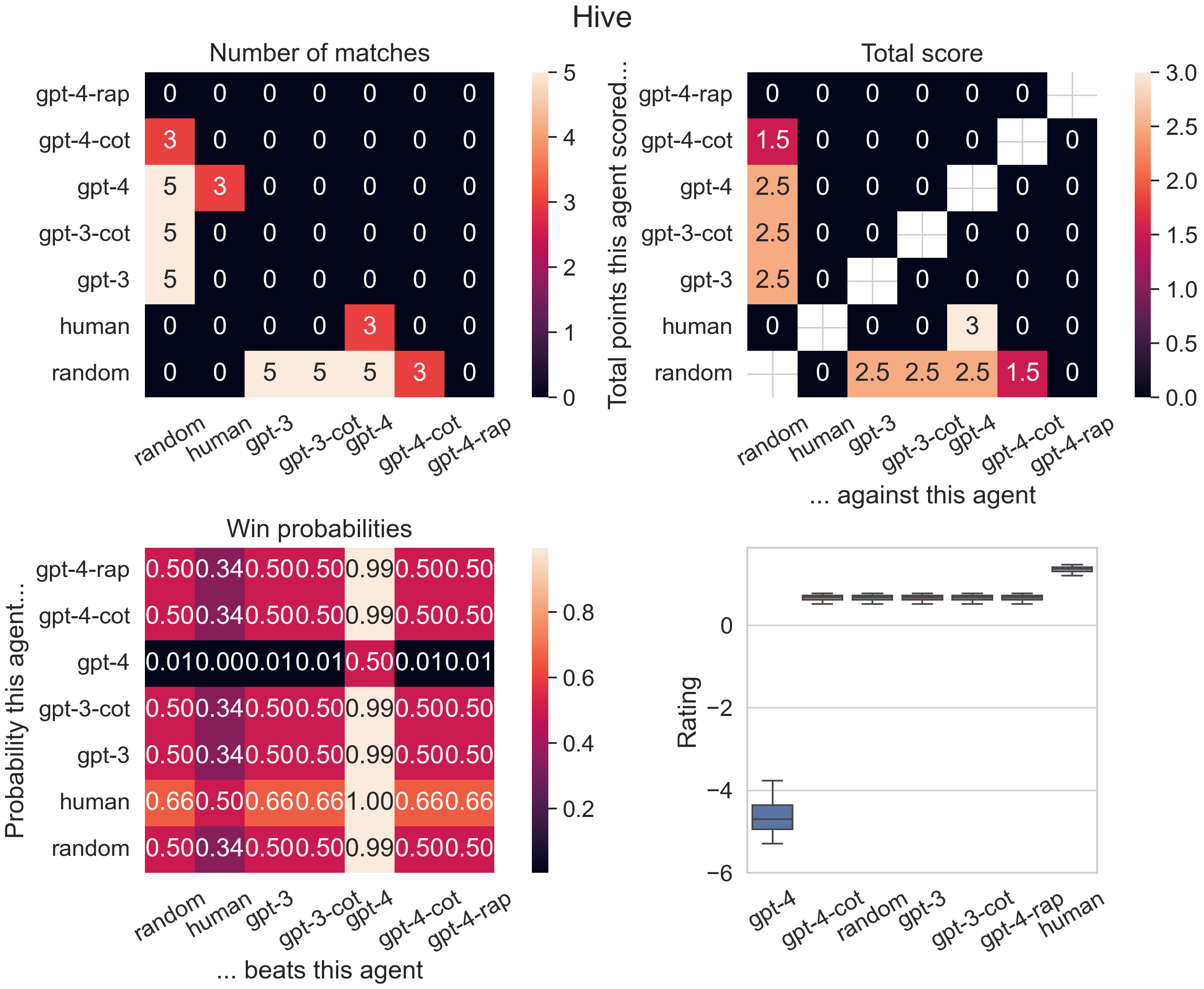}
\end{figure}
\begin{figure}
    \centering
    \includegraphics[width=0.9\textwidth]{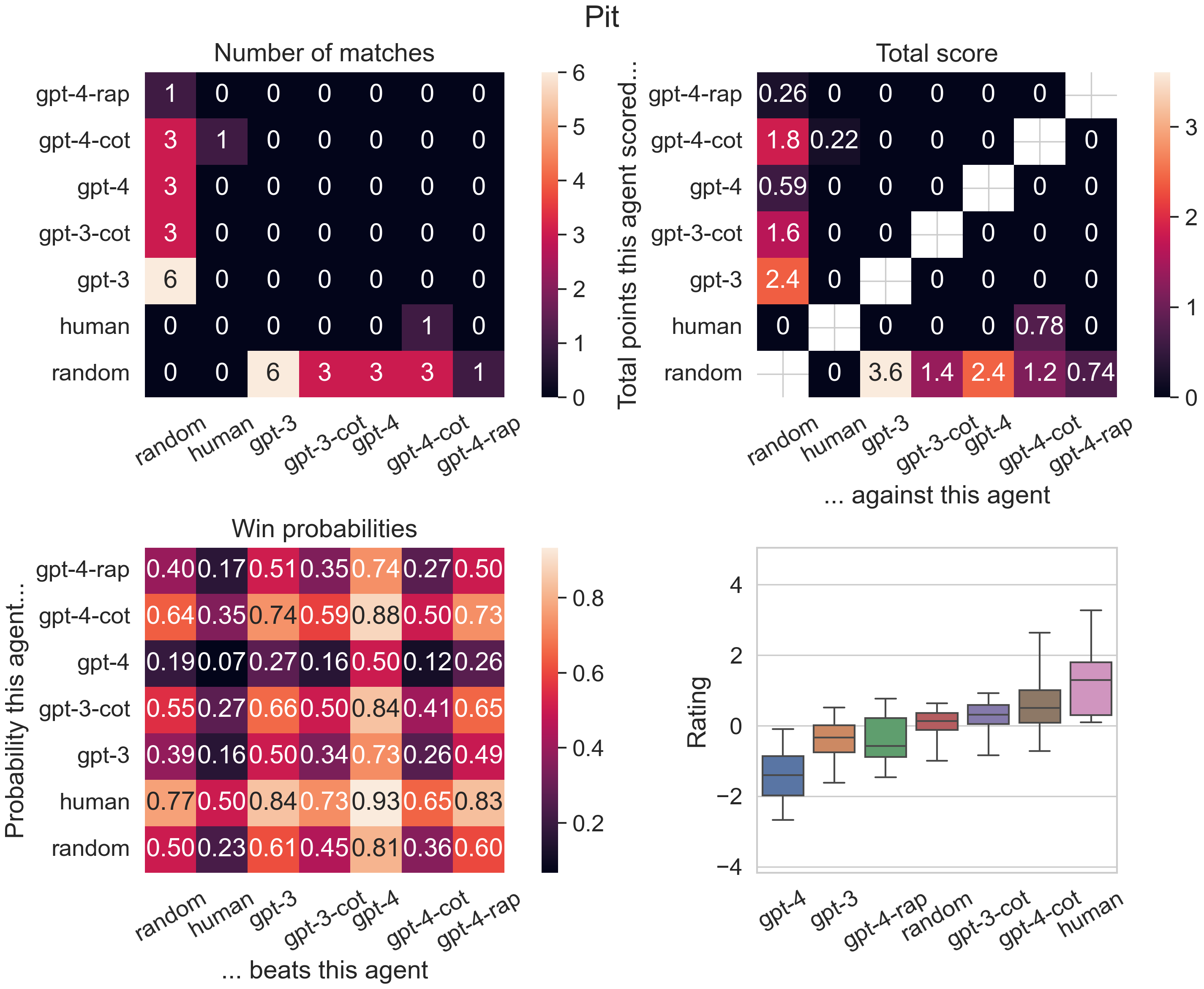}
\end{figure}
\begin{figure}
    \centering
    \includegraphics[width=0.9\textwidth]{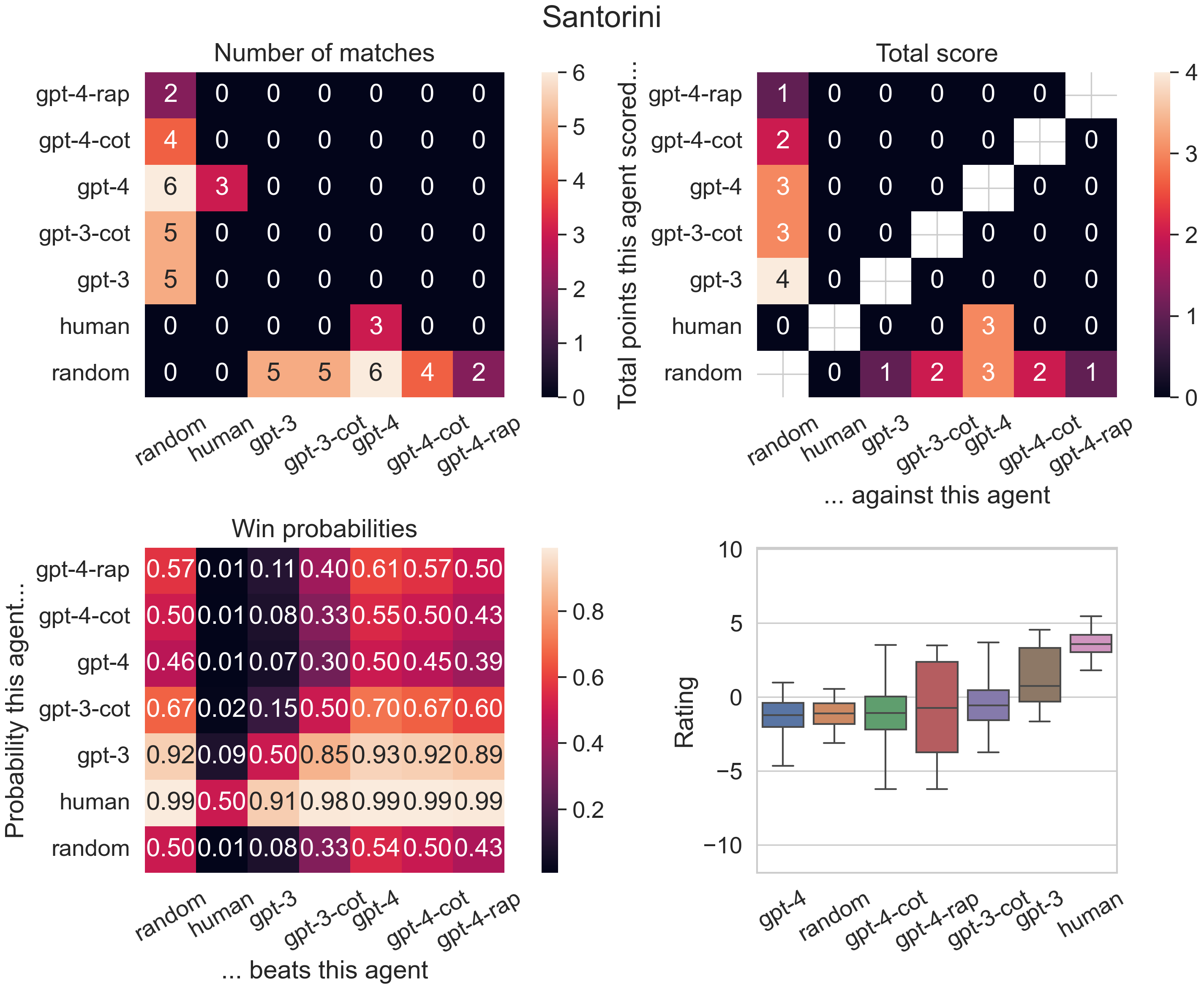}
\end{figure}
\begin{figure}
    \centering
    \includegraphics[width=0.9\textwidth]{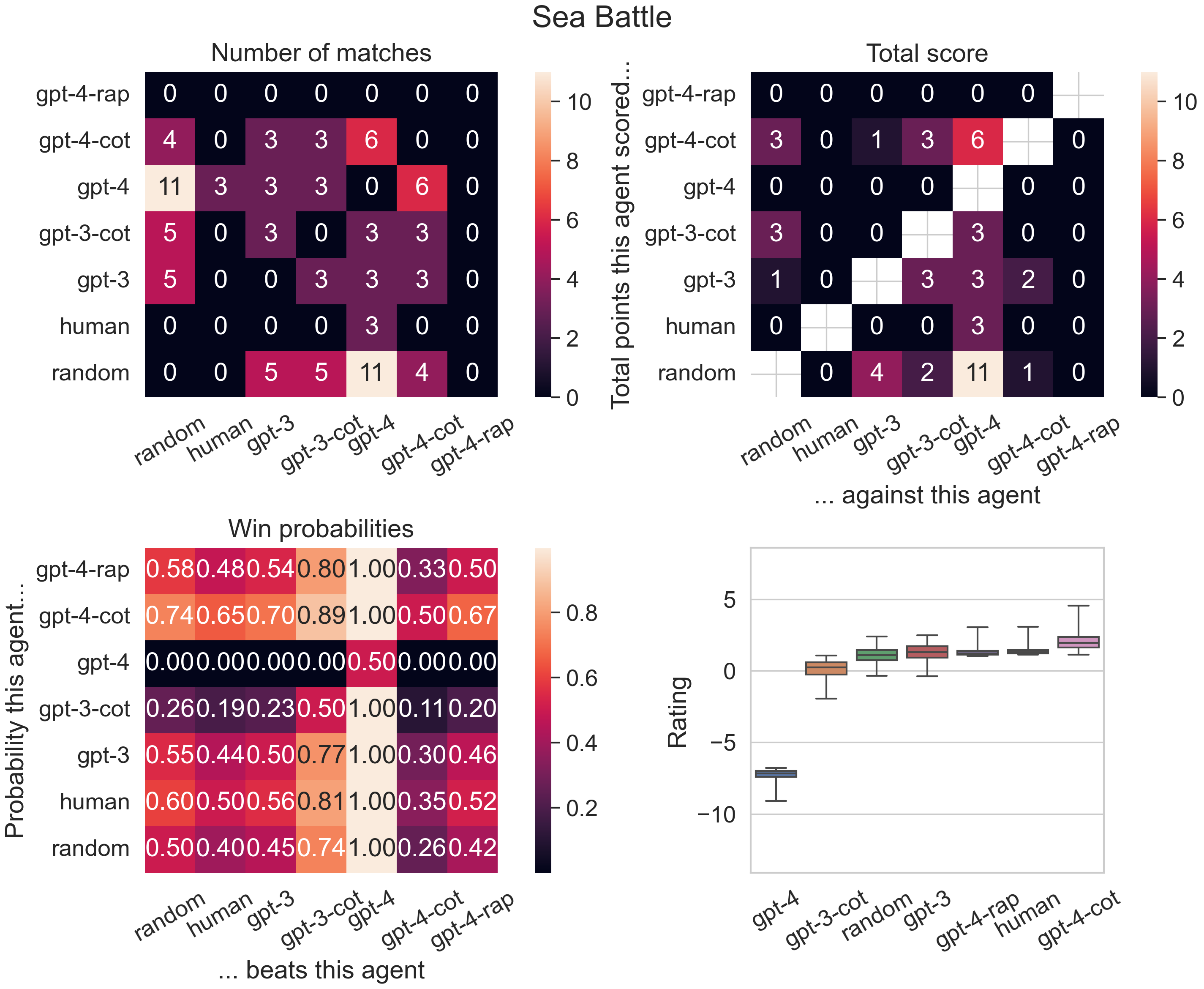}
\end{figure}
\begin{figure}
    \centering
    \includegraphics[width=0.9\textwidth]{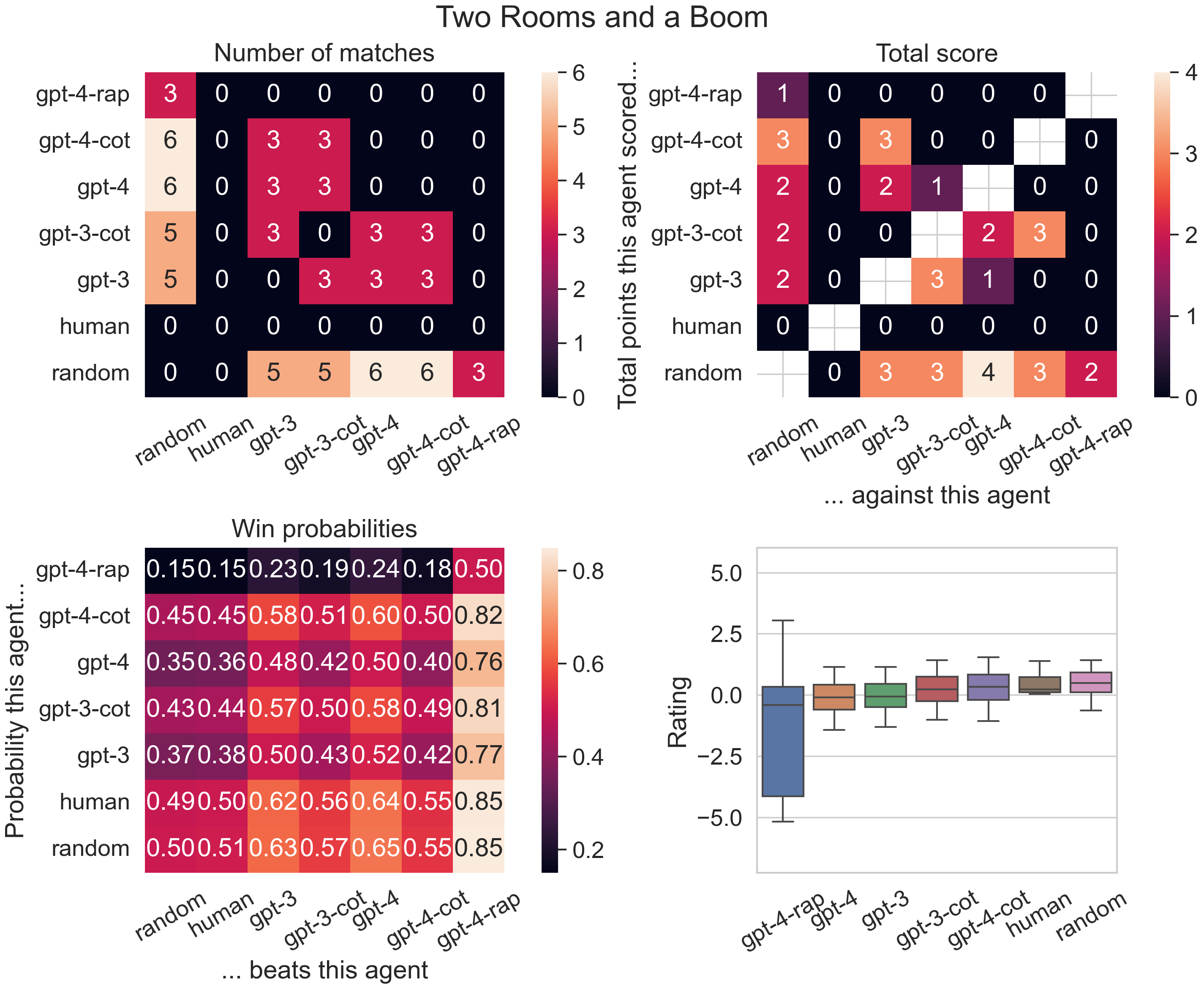}
\end{figure}

\end{document}